\documentclass[11pt,letterpaper]{article}
\usepackage{authblk} 
\usepackage{emnlp2016}
\usepackage{times}
\usepackage{latexsym}
\usepackage{graphicx}
\usepackage{amsmath}
\usepackage{amssymb}

\makeatletter
\newcommand{\@BIBLABEL}{\@emptybiblabel}
\newcommand{\@emptybiblabel}[1]{}
\makeatother
\usepackage[hidelinks]{hyperref}

\emnlpfinalcopy

\def\emnlppaperid{***}

\title{Towards a continuous modeling of natural language domains}

\author[1,2]{Sebastian Ruder}
\author[2]{Parsa Ghaffari}
\author[1]{John G. Breslin}
\affil[1]{Insight Centre for Data Analytics}
\affil[ ]{National University of Ireland, Galway}
\affil[ ]{\tt \{sebastian.ruder,john.breslin\}@insight-centre.org}
\affil[2]{Aylien Ltd.}
\affil[ ]{Dublin, Ireland}
\affil[ ]{\tt \{sebastian,parsa\}@aylien.com}

\date{}

\begin{document}

\maketitle

\begin{abstract}

Humans continuously adapt their style and language to a variety of domains. However, a reliable definition of `domain' has eluded researchers thus far. Additionally, the notion of discrete domains stands in contrast to the multiplicity of heterogeneous domains that humans navigate, many of which overlap. In order to better understand the change and variation of human language, we draw on research in domain adaptation and extend the notion of discrete domains to the continuous spectrum. We propose representation learning-based models that can adapt to continuous domains and detail how these can be used to investigate variation in language. To this end, we propose to use dialogue modeling as a test bed due to its proximity to language modeling and its social component.

\end{abstract}

\section{Introduction}
The notion of domain permeates natural language and human interaction: Humans continuously vary their language depending on the context, in writing, dialogue, and speech.
However, the concept of domain is ill-defined, with conflicting definitions aiming to capture the essence of what constitutes a domain. In semantics, a domain is considered a ``specific area of cultural emphasis'' \cite{Oppenheimer2006} that entails a particular terminology, e.g. a specific sport. In sociolinguistics, a domain consists of a group of related social situations, e.g. all human activities that take place at home. In discourse a domain is a ``cognitive construct (that is) created in response to a number of factors'' \cite{Douglas2004} and includes a variety of registers. Finally, in the context of transfer learning, a domain is defined as consisting of a feature space $\mathcal{X}$ and a marginal probability distribution $P(X)$ where $X = \{x_1, ..., x_n\}$ and $x_i$ is the $i^{th}$ feature vector \cite{Pan2010}. 

These definitions, although pertaining to different concepts, have a commonality: They separate the world in stationary domains that have clear boundaries. However, the real world is more ambiguous. Domains permeate each other and humans navigate these changes in domain.

Consequently, it seems only natural to step away from a \textit{discrete} notion of domain and adopt a \textit{continuous} notion. 
Utterances often cannot be naturally separated into discrete domains, but often arise from a continuous underlying process that is reflected in many facets of natural language: The web contains an exponentially growing amount of data, where each document ``is potentially its own domain'' \cite{McClosky2010}; a second-language learner adapts their style as their command of the language improves; language changes with time and with locality; even the WSJ section of the Penn Treebank -- often treated as a single domain -- contains different types of documents, such as news, lists of stock prices, etc. Continuity is also an element of real-world applications: In spam detection, spammers continuously change their tactics; in sentiment analysis, sentiment is dependent on trends emerging and falling out of favor.

Drawing on research in domain adaptation, we first compare the notion of continuous natural language domains against mixtures of discrete domains and motivate the choice of using dialogue modeling as a test bed. We then present a way of representing continuous domains and show how continuous domains can be incorporated into existing models. We finally propose a framework for evaluation.

\section{Continuous domains vs. mixtures of discrete domains}

In domain adaptation, a novel target domain is traditionally assumed to be discrete and independent of the source domain \cite{Blitzer2006}. Other research uses mixtures to model the target domain based on a single \cite{DaumeIII2006} or multiple discrete source domains \cite{Mansour2009a}. We argue that modeling a novel domain as a mixture of existing domains falls short in light of three factors.

Firstly, the diversity of human language makes it unfeasible to restrict oneself to a limited number of source domains, from which all target domains are modeled. This is exemplified by the diversity of the web, which contains billions of heterogeneous websites; the Yahoo! Directory\footnote{\url{https://en.wikipedia.org/wiki/Yahoo!_Directory}} famously contained thousands of hand-crafted categories in an attempt to separate these. Notably, many sub-categories were cross-linked as they could not be fully separated and websites often resided in multiple categories.

Similarly, wherever humans come together, the culmination of different profiles and interests gives rise to cliques, interest groups and niche communities that all demonstrate their own unique behaviors, unspoken rules, and memes. A mixture of existing domains fails to capture these varieties.

Secondly, using discrete domains for soft assignments relies on the assumption that the source domains are clearly defined. However, discrete labels only help to explain domains and make them interpretable, when in reality, a domain is a heterogeneous amalgam of texts. Indeed, Plank and van Noord \shortcite{Plank2011} show that selection based on human-assigned labels fares worse than using automatic domain similarity measures for parsing.

Thirdly, not only a speaker's style and command of a language are changing, but a language itself is continuously evolving. This is amplified in fast-moving media such as social platforms. Therefore, applying a discrete label to a domain merely anchors it in time. A probabilistic model of domains should in turn not be restricted to treat domains as independent points in a space. Rather, such a model should be able to walk the domain manifold and adapt to the underlying process that is producing the data. 

\section{Dialogue modeling as a test bed for investigating domains}

As a domain presupposes a social component and relies on context, we propose to use dialogue modeling as a test bed to gain a more nuanced understanding of how language varies with domain.

Dialogue modeling can be seen as a prototypical task in natural language processing akin to language modeling and should thus expose variations in the underlying language. It allows one to observe the impact of different strategies to model variation in language across domains on a downstream task, while being inherently unsupervised.

In addition, dialogue has been shown to exhibit characteristics that expose how language changes as conversation partners become more linguistically similar to each other over the course of the conversation \cite{Niederhoffer2002,Levitan2011}. Similarly, it has been shown that the linguistic patterns of individual users in online communities adapt to match those of the community they participate in \cite{Nguyen2011,Danescu-Niculescu-Mizil2013}.

For this reason, we have selected reddit as a medium and compiled a dataset from large amounts of reddit data. Reddit comments live in a rich environment that is dependent on a large number of contextual factors, such as community, user, conversation, etc. Similar to Chen et al. \shortcite{Chen2016a}, we would like to learn representations that allow us to disentangle factors that are normally intertwined, such as style and genre, and that will allow us to gain more insight about the variation in language. To this end, we are currently training models that condition on different communities, users, and threads. 

\section{Representing continuous domains}

\begin{figure}
	\centering
  	\includegraphics[width=\linewidth]{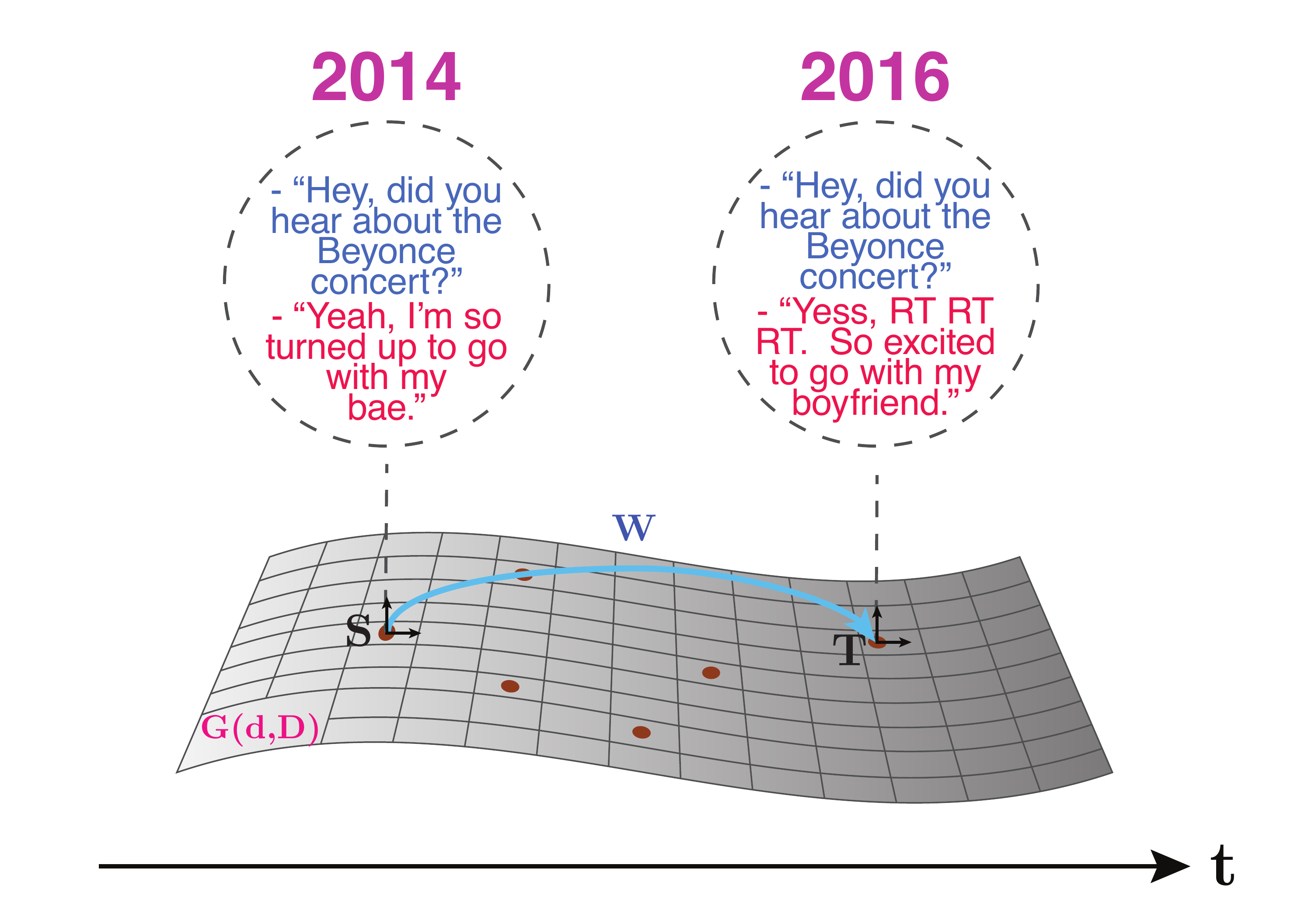}
  	\caption{Transforming a discrete source domain subspace $S$ into a target domain subspace $T$ with a transformation $W$.}
  	\label{fig:discrete}
\end{figure}

In line with past research \cite{DaumeIII2007a,Zhou2016}, we assume that every domain has an inherent low-dimensional structure, which allows its projection into a lower dimensional subspace. 

In the discrete setting, we are given two domains, a source domain $X_S$ and a target domain $X_T$. We represent examples in the source domain $X_S$ as $x_1^S, \cdots, x_{n_S}^S \in \mathbb{R}^d$ where $x_1^S$ is the $i$-th source example and $n_S$ is number of examples in $X_S$. Similarly, we have $n_T$ target domain examples $x_1^T, \cdots, x_{n_T}^T \in \mathbb{R}^d$.

We now seek to learn a transformation $W$ that allows us to transform the examples in the $X_S$ so that their distribution is more similar to the distribution of $X_T$. Equivalently, we can factorize the transformation $W$ into two transformations $A$ and $B$ with $W = AB^T$ that we can use to project the source and target examples into a joint subspace.

We assume that $X_S$ and $X_T$ lie on lower-dimensional orthonormal subspaces, $S, T \in \mathbb{R}^{D \times d}$, which can be represented as points on the Grassman manifold, $\mathcal{G}(d,D)$ as in Figure \ref{fig:discrete}, where $d \ll D$.

\begin{figure}
	\centering
  	\includegraphics[width=\linewidth]{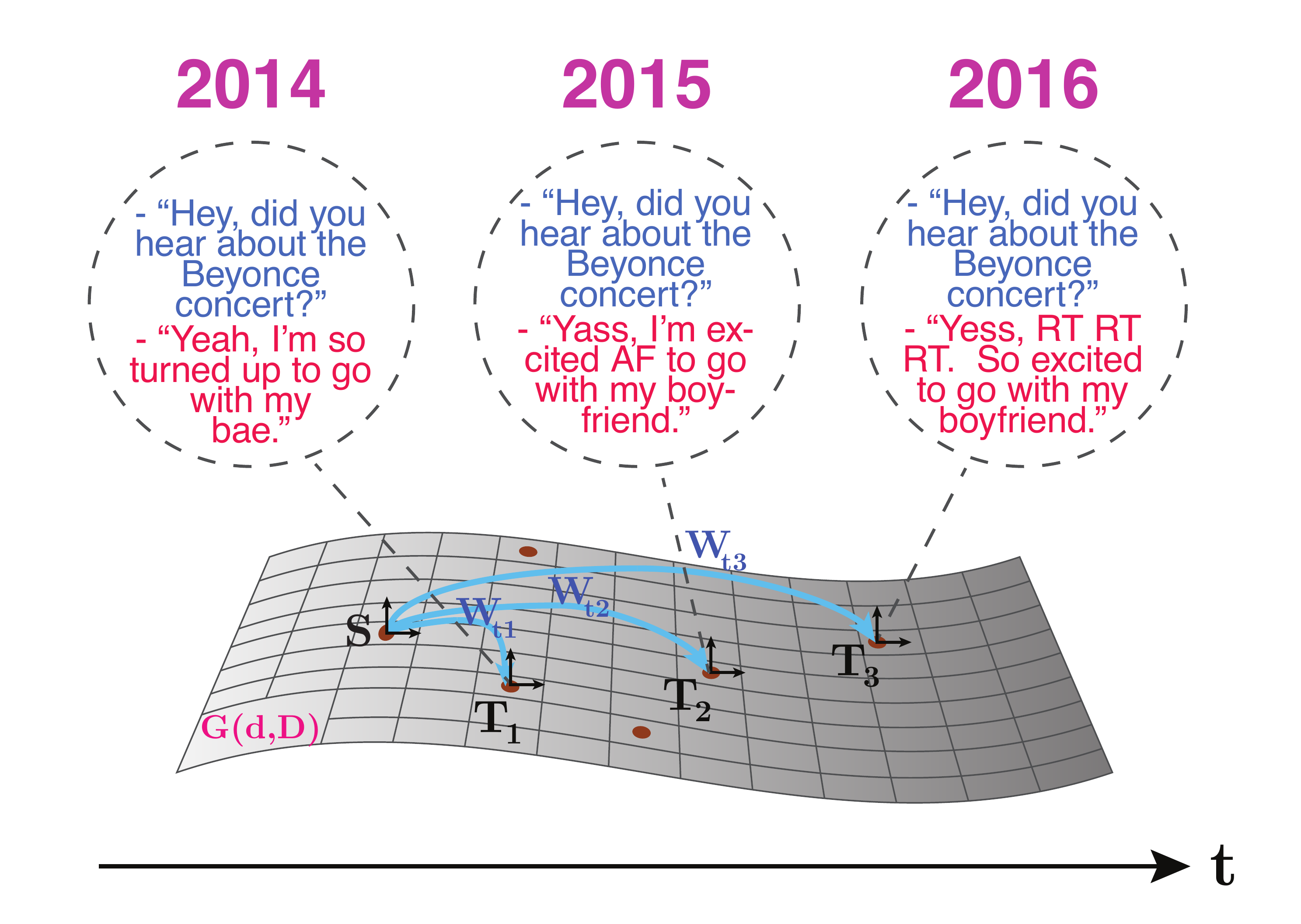}
  	\caption{Transforming a source domain subspace $S$ into continuous domain subspaces $T_t$ with a temporally varying transformation $W_t$.}
  	\label{fig:time}
\end{figure}

In computer vision, methods such as Subspace Alignment \cite{Fernando2013} or the Geodesic Flow Kernel \cite{Gong} have been used to find such transformations $A$ and $B$. Similarly, in natural language processing, CCA \cite{Faruqui2014} and Procrustes analysis \cite{Mogadala2016} have been used to align subspaces pertaining to different languages.

Many recent approaches using autoencoders \cite{Bousmalis2016,Zhou2016} learn such a transformation between discrete domains. Similarly, in a sequence-to-sequence dialogue model \cite{Vinyals2015a}, we can not only train the model to predict the source domain response, but also -- via a reconstruction loss -- its transformations to the target domain.

For continuous domains, we can assume that source domain $X_S$ and target domain $X_T$ are not independent, but that $X_T$ has evolved from $X_S$ based on a continuous process. This process can be indexed by time, e.g. in order to reflect how a language learner's style changes or how language varies as words rise and drop in popularity. We thus seek to learn a time-varying transformation $W_t$ between $S$ and $T$ that allows us to transform between source and target examples dependent on $t$ as in Figure \ref{fig:time}.

Hoffman et al. \shortcite{Hoffman2014} assume a stream of observations $z_1, \cdots, z_{n_t} \in \mathcal{R}^d$ drawn from a continuously changing domain and regularize $W_t$ by encouraging the new subspace at $t$ to be close to the previous subspace at $t-1$. Assuming a stream of (chronologically) ordered input data, a straightforward application of this to a representation-learning based dialogue model trains the parts of the model that auto-encode and transform the original message for each new example -- possibly regularized with a smoothness constraint -- while keeping the rest of the model fixed.

This can be seen as an unsupervised variant of fine-tuning, a common neural network domain adaptation baseline. As our learned transformation continuously evolves, we run the risk associated with fine-tuning of forgetting the knowledge acquired from the source domain. For this reason, neural network architectures that are immune to forgetting, such as the recently proposed Progressive Neural Networks \cite{Rusu} are appealing for continuous domain adaptation.

\begin{figure}
	\centering
  	\includegraphics[width=\linewidth]{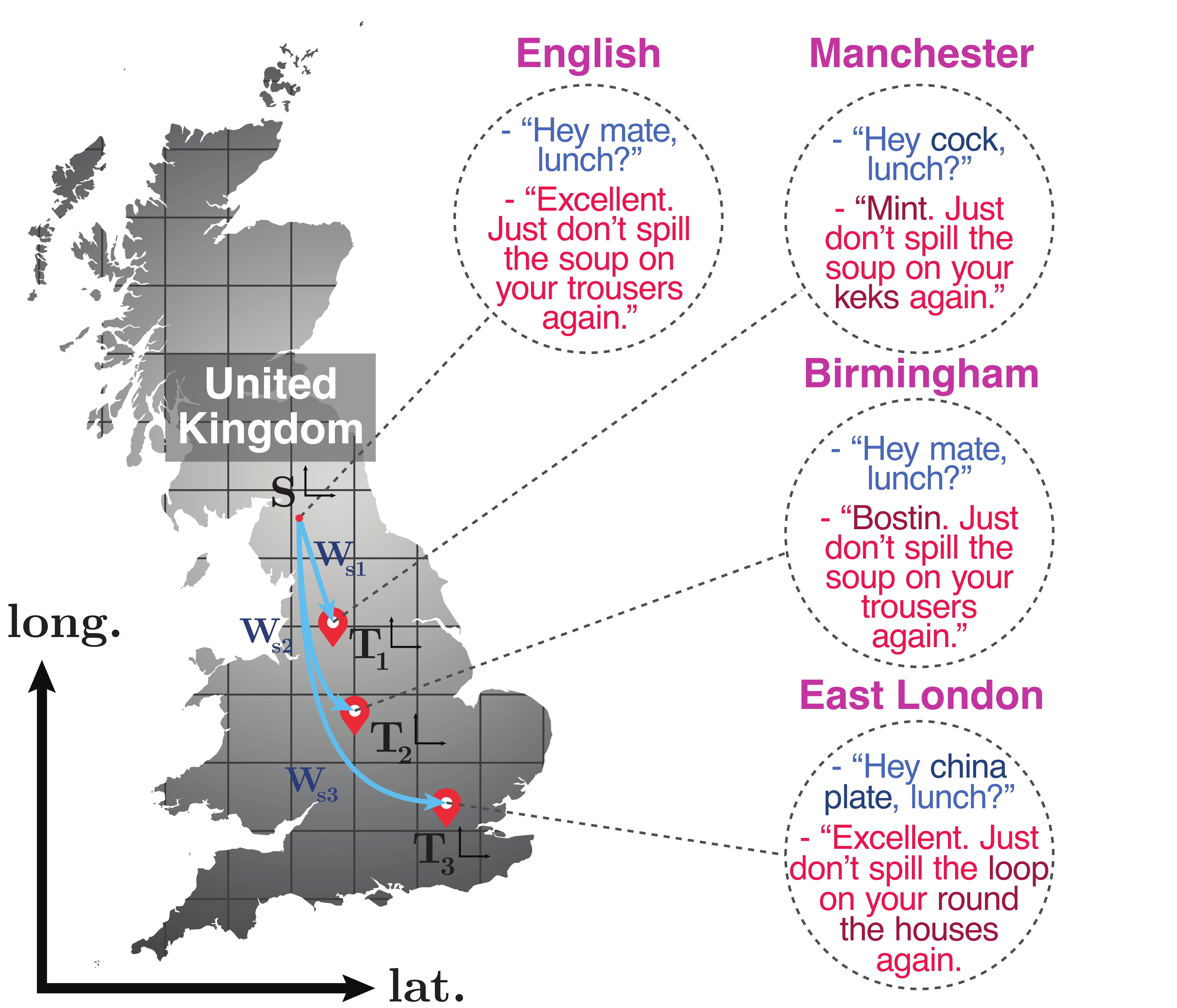}
  	\caption{Transforming a source domain subspace $S$ into continuous target domain subspaces $T_s$ using a spatially varying transformation $W_s$.}
  	\label{fig:space}
\end{figure}

While time is the most obvious dimension along which language evolves, other dimensions are possible: Geographical location influences dialectal variations as in Figure \ref{fig:space}; socio-economic status, political affiliation as well as a domain's purpose or complexity all influence language and can thus be conceived as axes that span a manifold for embedding domain subspaces.

\section{Investigating language change}

A continuous notion of domains naturally lends itself to a diachronic study of language. By looking at the representations produced by the model over different time steps, one gains insight into the change of language in a community or another domain. Similarly, observing how a user adapts their style to different users and communities reveals insights about the language of those entities.

Domain mixture models use various domain similarity measures to determine how similar the languages of two domains are, such as Renyi divergence \cite{VanAsch2010}, Kullback-Leibler (KL) divergence, Jensen-Shannon divergence, and vector similarity metrics \cite{Plank2011}, as well as task-specific measures \cite{Zhou2016}.

While word distributions have been used traditionally to compare domains, embedding domains in a manifold offers the possibility to evaluate the learned subspace representations. For this, cosine similarity as used for comparing word embeddings or KL divergence as used in the Variational Autoencoder \cite{Kingma2013} are a natural fit.

%
%

\section{Evaluation}

Our evaluation consists of three parts for evaluating the learned representations, the model, and the variation of language itself.

Firstly, as our models produce new representations for every subspace, we can compare a snapshot of a domain's representation after every $n$ time steps to chart a trajectory of its changes.

Secondly, as we are conducting experiments on dialogue modeling, gold data for evaluation is readily available in the form of the actual response. We can thus train a model on reddit data of a certain period, adapt it to a stream of future conversations and evaluate its performance with BLEU or another metric that might be more suitable to expose variation in language. At the same time, human evaluations will reveal whether the generated responses are faithful to the target domain.

Finally, the learned representations will allow us to investigate the variations in language. Ideally, we would like to walk the manifold and observe how language changes as we move from one domain to the other, similarly to \cite{Radford2015b}.

\section{Conclusion}

We have proposed a notion of continuous natural language domains along with dialogue modeling as a test bed. We have presented a representation of continuous domains and detailed how this representation can be incorporated into representation learning-based models. Finally, we have outlined how these models can be used to investigate change and variation in language. While our models allow us to shed light on how language changes, models that can adapt to continuous changes are key for personalization and the reality of grappling with an ever-changing world.


\bibliography{uphill_domain_adaptation}
\bibliographystyle{emnlp2016}

\end{document}